\documentclass{article}

\usepackage[final,nonatbib]{nips_2016}

\usepackage[utf8]{inputenc} % allow utf-8 input
\usepackage[T1]{fontenc}    % use 8-bit T1 fonts
\usepackage{url}            % simple URL typesetting
\usepackage{booktabs}       % professional-quality tables
\usepackage{amsfonts}       % blackboard math symbols
\usepackage{nicefrac}       % compact symbols for 1/2, etc.
\usepackage{microtype}      % microtypography
\usepackage{graphicx}
\usepackage{caption}
\usepackage{wrapfig,lipsum,booktabs}
\graphicspath{ {./} }

\newcommand{\etal}{et al.}

\usepackage{subfigure} 
\usepackage{amsmath}
\usepackage{amsthm}
\usepackage{algorithm}
\usepackage{algorithmic}
\usepackage{enumitem}

\title{Deep Motif Dashboard: Visualizing and Understanding Genomic Sequences Using Deep Neural Networks}

\author{
  Jack Lanchantin, Ritambhara Singh, Beilun Wang, and Yanjun Qi  \\
  Department of Computer Science\\
  University of Virginia\\
  Charlottesville, VA 22903 \\
  \texttt{\{jjl5sw,rs3zz,bw4mw,yq2h\}@virginia.edu} \\
}

\begin{document}

\maketitle

\begin{abstract}

Deep neural network (DNN) models have recently obtained state-of-the-art prediction accuracy for the transcription factor binding (TFBS) site classification task. However, it remains unclear how these approaches identify meaningful DNA sequence signals and give insights as to why TFs bind to certain locations. In this paper, we propose a toolkit called the Deep Motif Dashboard (DeMo Dashboard) which provides a suite of visualization strategies to extract motifs, or sequence patterns from deep neural network models for TFBS classification. We demonstrate how to visualize and understand three important DNN models: convolutional, recurrent, and convolutional-recurrent networks. Our first visualization method is finding a test sequence's saliency map which uses first-order derivatives to describe the importance of each nucleotide in making the final prediction. Second, considering recurrent models make predictions in a temporal manner (from one end of a TFBS sequence to the other), we introduce temporal output scores, indicating the prediction score of a model over time for a sequential input. Lastly, a class-specific visualization strategy finds the optimal input sequence for a given TFBS positive class via stochastic gradient optimization. Our experimental results indicate that a convolutional-recurrent architecture performs the best among the three architectures. The visualization techniques indicate that CNN-RNN makes predictions by modeling both motifs as well as dependencies among them.

\end{abstract}

\section{Introduction}

In recent years, there has been an explosion of deep learning models which have lead to groundbreaking results in many fields such as computer vision\cite{krizhevsky2012imagenet}, natural language processing\cite{sutskever2014sequence}, and computational biology \cite{alipanahi2015predicting,quang2015danq,zhou2015predicting,kelley2016basset,lanchantin2016motif, Singh01092016}. However, although these models have proven to be very accurate, they have widely been viewed as ``black boxes'' due to their complexity, making them hard to understand. This is particularly unfavorable in the biomedical domain, where understanding a model's predictions is extremely important for doctors and researchers trying to use the model. 

Aiming to open up the black box, we present the ``Deep Motif Dashboard\footnote{Dashboard normally refers to a user interface that gives a current summary, usually in graphic, easy-to-read form, of key information relating to performance\cite{dashboard}.}'' (DeMo Dashboard), to understand the inner workings of deep neural network models for a genomic sequence classification task. We do this by introducing a suite of different neural models and visualization strategies to see which ones perform the best and understand how they make their predictions.\footnote{We implemented our model in Torch, and it is made available at deepmotif.org}

Understanding genetic sequences is one of the fundamental tasks of health advancements due to the high correlation of genes with diseases and drugs. An important problem within genetic sequence understanding is related to transcription factors (TFs), which are regulatory proteins that bind to DNA. Each different TF binds to specific transcription factor binding sites (TFBSs) on the genome to regulate cell machinery. Given an input DNA sequence, classifying whether or not there is a binding site for a particular TF is a core task of bioinformatics\cite{stormo2000dna}. 

%Although different TFs bind in different locations on DNA simultaneously, we focus on predicting and understanding where a particular TF binds independently of other TFs.

For our task, we follow a two step approach. First, given a particular TF of interest and a dataset containing samples of positive and negative TFBS sequences, we construct three deep learning architectures to classify the sequences. Section \ref{method} introduces the three different DNN structures that we use: a convolutional neural network (\textbf{CNN}), a recurrent neural network (\textbf{RNN}), and a convolutional-recurrent neural network (\textbf{CNN-RNN}).

Once we have our trained models to predict binding sites, the second step of our approach is to understand why the models perform the way they do. As explained in section \ref{dgd_method}, we do this by introducing three different visualization strategies for interpreting the models:\vspace{-1ex}
\begin{enumerate}
  \item Measuring nucleotide importance with \textbf{Saliency Maps}.
  \item Measuring critical sequence positions for the classifier using \textbf{Temporal Output Scores}.
  \item Generating class-specific motif patterns with \textbf{Class Optimization}.
\end{enumerate}\vspace{-0.9ex}

We test and evaluate our models and visualization strategies on a large scale benchmark TFBS dataset. Section \ref{experiments} provides experimental results for understanding and visualizing the three DNN architectures. We find that the CNN-RNN outperforms the other models. From the visualizations, we observe that the CNN-RNN tends to focus its predictions on the traditional motifs, as well as modeling long range dependencies among motifs.

\section{Deep Neural Models for TFBS Classification}
\label{method}
\paragraph{TFBS Classification.\,} 
Chromatin immunoprecipitation (ChIP-seq) technologies and databases such as ENCODE \cite{encode2012integrated} have made binding site locations available for hundreds of different TFs. Despite these advancements, there are two major drawbacks: (1) ChIP-seq experiments are slow and expensive, (2) although ChIP-seq experiments can find the binding site locations, they cannot find patterns that are common across all of the positive binding sites which can give insight as to why TFs bind to those locations. Thus, there is a need for large scale computational methods that can not only make accurate binding site classifications, but also identify and understand patterns that influence the binding site locations.

In order to computationally predict TFBSs on a DNA sequence, researchers initially used consensus sequences and position weight matrices to match against a test sequence \cite{stormo2000dna}. Simple neural network classifiers were then proposed to differentiate positive and negative binding sites, but did not show significant improvements over the weight matrix matching methods \cite{horton1992assessment}. Later, SVM techniques outperformed the generative methods by using k-mer features \cite{ghandi2014enhanced, setty2015seqgl}, but string kernel based SVM systems are limited by expensive computational cost proportional to the number of training and testing sequences. Most recently, convolutional neural network models have shown state-of-the-art results on the TFBS task and are scalable to a large number of genomic sequences \cite{alipanahi2015predicting,lanchantin2016motif}, but it remains unclear which neural architectures work best.

\begin{figure}[tp]
  \centering
\includegraphics[width=15cm,height=11cm]{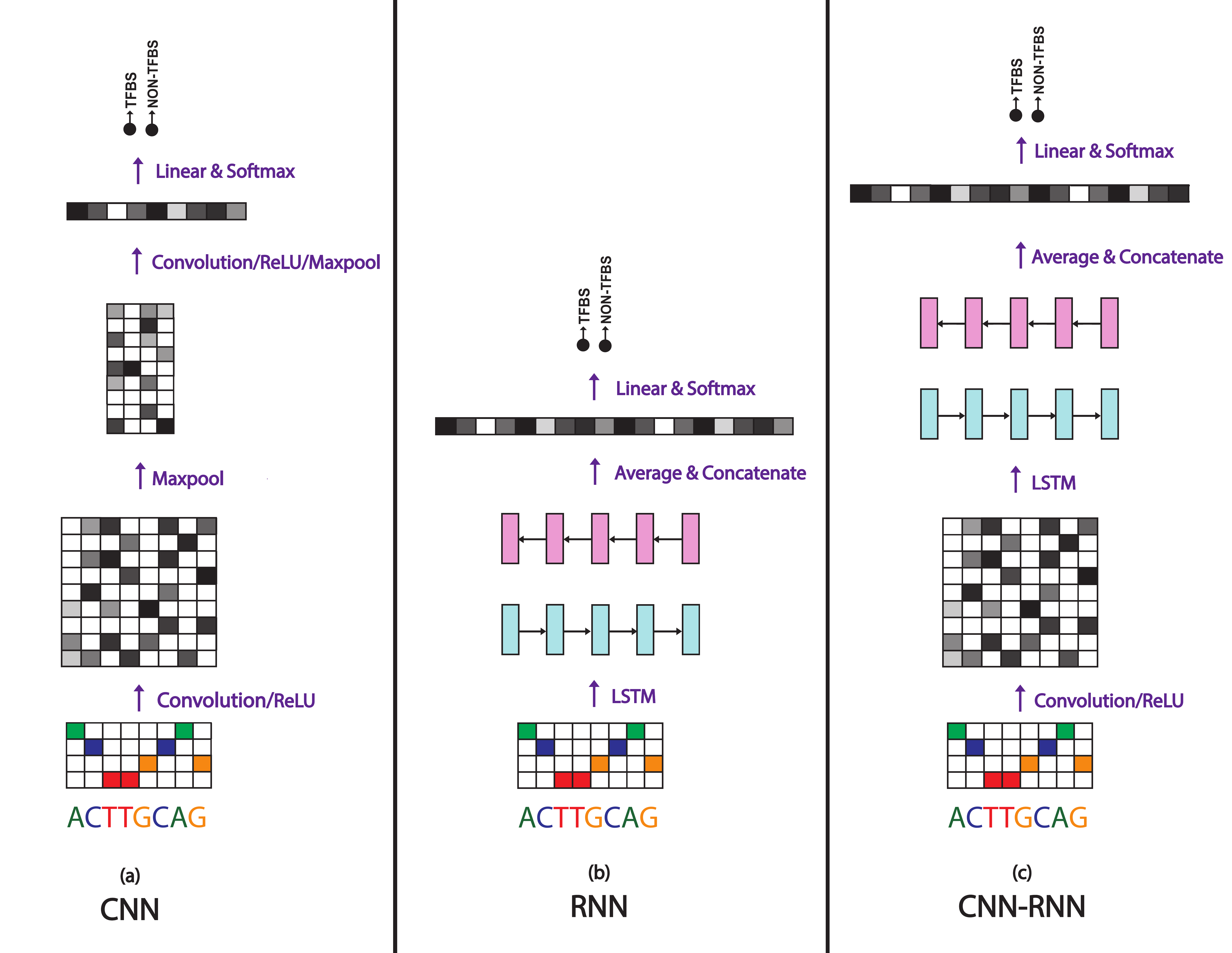}
\caption{\textbf{Model Architectures.} Each model has the same input (one-hot encoded matrix of the raw nucleotide inputs), and the same output (softmax classifier to make a binary prediction). The architectures differ by the middle ``module'', which are \textbf{(a)} Convolutional,  \textbf{(b)} Recurrent, and \textbf{(c)} Convolutional-Recurrent.}
\label{fig:models}
\end{figure}

\paragraph{Deep Neural Networks for TFBSs.\,}
To find which neural models work the best on the TFBS classification task, we examine several different types of models. Inspired by their success across different fields, we explore variations of two popular deep learning architectures: convolutional neural networks (CNNs), and recurrent neural networks (RNNs). CNNs have dominated the field of computer vision in recent years, obtaining state-of-the-art results in many tasks due to their ability to automatically extract translation-invariant features. On the other hand, RNNs have emerged as one of the most powerful models for sequential data tasks such as natural language processing due to their ability to learn long range dependencies. Specifically, on the TFBS prediction task, we explore three distinct architectures: (1) CNN, (2) RNN, and (3) a combination of the two, CNN-RNN. Figure \ref{fig:models} shows an overview of the models.

%For example, when doing face detection, they can extract features from an image such as an eye, invariant to the location or translation of the eye in the image.
%Different from ordinary neural networks, the connections between RNN units feed back to themselves, allowing the model to learn long range dependencies.

\paragraph{End-to-end Deep Framework.\,}
While the body of the three architectures we use differ, each implemented model follows a similar end-to-end framework which we use to easily compare and contrast results. We use the raw nucleotide characters (A,C,G,T) as inputs, where each character is converted into a one-hot encoding (a binary vector with the matching character entry being a $1$ and the rest as $0$s). This encoding matrix is used as the input to a convolutional, recurrent, or convolutional-recurrent module that each outputs a vector of fixed dimension. The output vector of each model is linearly fed to a softmax function as the last layer which learns the mapping from the hidden space to the output class label space $C \in [+1,-1]$.  The final output is a probability indicating whether an input is a positive or a negative binding site (binary classification task). The parameters of the network are trained end-to-end by minimizing the negative log-likelihood over the training set. The minimization of the loss function is obtained via the stochastic gradient algorithm Adam\cite{kingma2014adam}, with a mini-batch size of 256 sequences. We use dropout \cite{srivastava2014dropout} as a regularization method for each model. 

%In order for us to classify into class $c\in C$ for sequence $x$, we apply a softmax
%operator on the outputs $f_{c,x}$. Given the parameters of the network $\theta$, this gives us a conditional probability of class $c$:
%\begin{equation} 
%p(c\in C|f_{c,x},\theta) = \frac{e^{f_{c,x}}}{\sum_{c \in C} e^{f_{c,x}}}.
%\end{equation}
%\begin{equation} 
%L(\theta) = -\sum_{x\in S}{\ln p(c_{correct} | f_{c,x}, \theta)},
%\end{equation}
%where, $f_{c,x}$ represents the softmax output, $c_{correct}$ is the correct TFBS label of sequence $x$. 

\setlength{\textfloatsep}{10pt plus 1.0pt minus 1.0pt}
\setlength{\textfloatsep}{10pt plus 1.0pt minus 1.0pt}

\subsection{Convolutional Neural Network (CNN)}

In genomic sequences, it is believed that regulatory mechanisms such as transcription factor binding are influenced by local sequential patterns known as ``motifs''. Motifs can be viewed as the temporal equivalent of spatial patterns in images such as eyes on a face, which is what CNNs are able to automatically learn and achieve state-of-the art results on computer vision tasks. As a result, a temporal convolutional neural network is a fitting model to automatically extract these motifs. A temporal convolution with filter (or kernel) size $k$ takes an input data matrix $\mathbf{X}$ of size $T\times n_{in}$, with length $T$  and input layer size $n_{in}$, and outputs a matrix $\mathbf{Z}$ of size $T\times n_{out}$, where $n_{out}$ is the output layer size. Specifically, $convolution(\mathbf{X}) = \mathbf{Z}$, where \vspace{-0.1in}
\begin{equation}  
\mathbf{z}_{t,i} = \sigma(\mathbf{B}_i + \sum_{j=1}^{n_{in}} \sum_{z=1}^{k} \mathbf{W}_{i,j,z} \mathbf{x}_{t+z-1,j}),
\end{equation} 
where $\mathbf{W}$ and $\mathbf{B}$ are the trainable parameters of the convolution filter, and $\sigma$ is a function enforcing element-wise nonlinearity. We use rectified linear units (ReLU) as the nonlinearity: 
\begin{equation}
\mathrm{ReLU}(x) = \max(0, x).
\end{equation}\vspace{-0.2in}

After the convolution and nonlinearity, CNNs typically use maxpooling, which is a dimension reduction technique to provide translation invariance and to extract higher level features from a wider range of the input sequence. Temporal maxpooling on a matrix $\mathbf{Z}$ with a pooling size of $m$ results in output matrix $\mathbf{Y}$. Formally, $maxpool(\mathbf{Z}) = \mathbf{Y}$, where
\begin{equation} 
\mathbf{y}_{t,i} = \max_{j=1}^m { \mathbf{z}_{m(t-1)+j, i} }
\end{equation}

Our CNN implementation involves a progression of convolution, nonlinearity, and maxpooling. This is represented as one convolutional layer in the network, and we test up to 4 layer deep CNNs. The final layer involves a maxpool across the entire temporal domain so that we have a fixed-size vector which can be fed into a softmax classifier. 

Figure \ref{fig:models} (a) shows our CNN model with two convolutional layers. The input one-hot encoded matrix is convolved with several filters (not shown) and fed through a ReLU nonlinearity to produce a matrix of convolution activations. We then perform a maxpool on the activation matrix. The output of the first maxpool is fed through another convolution, ReLU, and maxpooled across the entire length resulting in a vector. This vector is then transposed and fed through a linear and softmax layer for classification.

\subsection{Recurrent Neural Network (RNN)}
Designed to handle sequential data, Recurrent neural networks (RNNs) have become the main neural model for tasks such as natural language understanding. The key advantage of RNNs over CNNs is that they are able to find long range patterns in the data which are highly dependent on the ordering of the sequence for the prediction task.

Given an input matrix $\mathbf{X}$ of size $T\times n_{in}$, an RNN produces matrix $\mathbf{H}$ of size $T\times d$, where $d$ is the RNN embedding size. At each timestep $t$, an RNN takes an input column vector $\mathbf{x_t} \in \mathbb{R}^{n_{in}}$ and the previous hidden state vector $\mathbf{h_{t-1}} \in \mathbb{R}^{d}$ and produces the next hidden state $\mathbf{h_{t}}$ by applying the following recursive operation:
\begin{equation}
\mathbf{h}_{t} = \sigma(\mathbf{W}x_{t }+ \mathbf{Uh}_{t-1} + \mathbf{b}),
\end{equation}
where $\mathbf{W}, \mathbf{U}, \mathbf{b}$ are the trainable parameters of the model, and $\sigma$ is an element-wise nonlinearity. Due to their recursive nature, RNNs can model the full conditional distribution of any sequential data and find dependencies over time, where each position in a sequence is a timestep on an imaginary time coordinate running in a certain direction. To handle the ``vanishing gradients'' problem of training basic RNNs on long sequences, Hochreiter and Schmidhuber \cite{hochreiter1997long} proposed an RNN variant called the Long Short-term Memory (LSTM) network (for simplicity, we refer to LSTMs as RNNs in this paper), which can handle long term dependencies by using gating functions. These gates can control when information is written to, read from, and forgotten. Specifically, LSTM ``cells'' take inputs $\mathbf{x}_t, \mathbf{h}_{t-1} $, and $\mathbf{c}_{t-1} $, and produce $\mathbf{h}_{t}$, and $\mathbf{c}_{t}$:
\begin{align*}
\mathbf{i}_t &= \sigma(\mathbf{W}^i\mathbf{x}_t + \mathbf{U}^i\mathbf{h}_{t-1} + \mathbf{b}^i) \\
\mathbf{f}_t &= \sigma(\mathbf{W}^f\mathbf{x}_t + \mathbf{U}^f\mathbf{h}_{t-1} + \mathbf{b}^f) \\
\mathbf{o}_t &= \sigma(\mathbf{W}^o\mathbf{x}_t + \mathbf{U}^o\mathbf{h}_{t-1} + \mathbf{b}^o) \\
\mathbf{g}_t &= tanh(\mathbf{W}^g\mathbf{x}_t + \mathbf{U}^g\mathbf{h}_{t-1} + \mathbf{b}^g) \\
\mathbf{c}_t &= \mathbf{f}_t \odot \mathbf{c}_{t-1} + \mathbf{i}_t \odot \mathbf{g}_t \\
\mathbf{h}_t &= \mathbf{o}_t \odot tanh(\mathbf{c}_t)
\end{align*}
where $\sigma(\cdot)$, $tanh(\cdot)$, and $\odot$ are element-wise sigmoid, hyperbolic tangent, and multiplication functions, respectively. $\mathbf{i}_t$, $\mathbf{f}_t$, and $\mathbf{o}_t$ are the input, forget, and output gates, respectively.

RNNs produce an output vector $\mathbf{h}_t$ at each timestep $t$ of the input sequence. In order to use them on a classification task, we take the mean of all vectors $\mathbf{h}_t$, and use the mean vector $\mathbf{h}_{mean} \in \mathbb{R}^d$ as input to the softmax layer.
%\begin{equation}
%\mathbf{m} = \textrm{mean}(\{\mathbf{h}_1,\mathbf{h}_2,...,\mathbf{h}_t\})
%\end{equation}
%The vector $\bf{m}$ $\in 

Since there is no innate direction in genomic sequences, we use a bi-directional LSTM as our RNN model. In the bi-directional LSTM, the input sequence gets fed through two LSTM networks, one in each direction, and then the output vectors of each direction get concatenated together in the temporal direction and fed through a linear classifier.

Figure \ref{fig:models} (b) shows our RNN model. The input one-hot encoded matrix is fed through an LSTM in both the forward and backward direction which each produce a matrix of column vectors representing the LSTM output embedding at each timestep. These vectors are then averaged to create one vector for each direction representing the LSTM output. The forward and backward output vectors are then concatenated and fed to the softmax for classification.

\subsection{Convolutional-Recurrent Network (CNN-RNN)}
Considering convolutional networks are designed to extract motifs, and recurrent networks are designed to extract temporal features, we implement a combination of the two in order to find temporal patterns between the motifs. Given an input matrix $\mathbf{X} \in \mathbb{R}^{T\times n_{in}}$, the output of the CNN is $\mathbf{Z} \in \mathbb{R}^{T\times n_{out}}$. Each column vector of $\mathbf{Z}$ gets fed into the RNN one at a time in the same way that the one-hot encoded vectors get input to the regular RNN model. The resulting output of the RNN $\mathbf{H} \in \mathbb{R}^{T\times d}$, where $d$ is the LSTM embedding size, is then averaged across the temporal domain (in the same way as the regular RNN), and fed to a softmax classifier.

Figure \ref{fig:models} (c) shows our CNN-RNN model. The input one-hot encoded matrix is fed through one layer of convolution to produce a convolution activation matrix. This matrix is then input to the LSTM, as done in the regular RNN model from the original one-hot matrix. The output of the LSTM is averaged, concatenated, and fed to the softmax, similar to the RNN.

\section{Visualizing and Understanding Deep Models}
\label{dgd_method}
The previous section explained the deep models we use for the TFBS classification task, where we can evaluate which models perform the best. While making accurate predictions is important in biomedical tasks, it is equally important to understand why models make their predictions. Accurate, but uninterpretable models are often very slow to emerge in practice due to the inability to understand their predictions, making biomedical domain experts reluctant to use them. Consequently, we aim to obtain a better understanding of why certain models work better than others, and investigate how they make their predictions by introducing several visualization techniques. The proposed DeMo Dashboard allows us visualize and understand DNNs in three different ways: Saliency Maps, Temporal Output Scores, and Class Optimizations.

%TODO: need this?
%The visualization strategies we explore in the DeMo Dashboard are inspired by recent work attempting to understand the deep models in computer vision and natural language processing. For example, given a deep network trained to classify objects in an image, trying to find patterns in the network which correspond to how humans would classify such image \cite{simonyan2013deep,zeiler2014visualizing}. Similarly, given a neural network trained to generate natural language, trying to find interpretable neurons of the network which computationally follow known language semantics \cite{karpathy2015visualizing}. The main difference in our work is that instead of trying to understand the neural models given human understanding of such human perception tasks, the we attempt to uncover critical signals in DNA sequences given the deep models.

\subsection{Saliency Maps}
For a certain DNA sequence and a model's classification, a logical question may be: ``which which parts of the sequence are most influential for the classification?'' To do this, we seek to visualize the influence of each position (i.e. nucleotide) on the prediction. Our approach is similar to the methods used on images by Simonyan \etal \cite{simonyan2013deep} and Baehrens \etal\cite{baehrens2010explain}. Given a sequence $X_0$ of length $|X_0|$, and class $c \in C$, a DNN model provides a score function $S_c(X_0)$. We rank the nucleotides of $X_0$ based on their influence on the score $S_c(X_0)$. Since $S_c(X)$ is a highly non-linear function of $X$ with deep neural nets, it is hard to directly see the influence of each nucleotide of $X$ on $S_c$. Mathematically, around the point $X_0$, $S_c(X)$ can be approximated by a linear function by computing the first-order Taylor expansion:
\begin{equation}
\label{saliencymapeq}
S_c(X) \approx w^TX + b = \sum_{i=1}^{|X|} w_ix_i + b
\end{equation}
where $w$ is the derivative of $S_c$ with respect to the sequence variable $X$ at the point $X_0$:
\begin{equation}
w = \frac{\partial S_c}{\partial X} \bigg|_{X_0} =\, saliency\,\, map
\end{equation}
	This derivative is simply one step of backpropagation in the  DNN model, and is therefore easy to compute. We do a pointwise multiplication of the saliency map with the one-hot encoded sequence to get the derivative values for the actual nucleotide characters of the sequence (A,T,C, or G) so we can see the influence of the character at each position on the output score. Finally, we take the element-wise magnitude of the resulting derivative vector to visualize how important each character is regardless of derivative direction. We call the resulting vector a ``saliency map\cite{simonyan2013deep}'' because it tells us which nucleotides need to be changed the least in order to affect the class score the most. As we can see from equation \ref{saliencymapeq}, the saliency map is simply a weighted sum of the input nucleotides, where the each weight, $w_i$, indicates the influence of that nucleotide position on the output score.

\subsection{Temporal Output Scores}
Since DNA is sequential (i.e. can be read in a certain direction), it can be insightful to visualize the output scores at each timestep (position) of a sequence, which we call the temporal output scores. Here we assume an imaginary time direction running from left to right on a given sequence, so each position in the sequence is a timestep in such an imagined time coordinate. In other words, we check the RNN's prediction scores when we vary the input of the RNN. The input series is constructed by using subsequences of an input $X$ running along the imaginary time coordinate, where the subsequences start from just the first nucleotide (position), and ends with the entire sequence $X$. This way we can see exactly where in the sequence the recurrent model changes its decision from negative to positive, or vice versa. Since our recurrent models are bi-directional, we also use the same technique on the reverse sequence. CNNs process the entire sequence at once, thus we can't view its output as a temporal sequence, so we use this visualization on just the RNN and CNN-RNN.

\subsection{Class Optimization}
The previous two visualization methods listed are representative of a specific testing sample (i.e. sequence-specific). Now we introduce an approach to extract a \textit{class-specific} visualization for a DNN model, where we attempt to find the best sequence which maximizes the probability of a positive TFBS, which we call class optimization. Formally, we optimize the following equation where $S_{+}(X)$ is the probability (or score) of an input sequence $X$ (matrix in our case) being a positive TFBS computed by the softmax equation of our trained DNN model for a specific TF: 
\abovedisplayskip=5pt
\abovedisplayshortskip=5pt
\belowdisplayskip=5pt
\belowdisplayshortskip=5pt
\begin{equation}
\arg\max\limits_{X} S_{+}(X) + \lambda \|X\|_2^2
\label{eq:1}
\end{equation}
where $\lambda$ is the regularization parameter. We find a locally optimal $X$ through stochastic gradient descent, where the optimization is with respect to the input sequence. In this optimization, the model weights remain unchanged. This is similar to the methods used in Simonyan \etal \cite{simonyan2013deep} to optimize toward a specific image class. This visualization method depicts the notion of a positive TFBS class for a particular TF and is not specific to any test sequence. 

\subsection{End-to-end Automatic Motif Extraction from the Dashboard\,} 
\label{e2emotif}
Our three proposed visualization techniques allow us to manually inspect how the models make their predictions. In order to automatically find patterns from the techniques, we also propose methods to extract motifs, or consensus subsequences that represent the positive binding sites. We extract motifs from each of our three visualization methods in the following ways: (1) From each positive test sequence (thus, 500 total for each TF dataset) we extract a motif from the saliency map by selecting the contiguous length-9 subsequence that achieves the highest sum of contiguous length-9 saliency map values.  (2) For each positive test sequence, we extract a motif from the temporal output scores by selecting the length-9 subsequence that shows the strongest score change from negative to positive output score. (3) For each different TF, we can directly use the class-optimized sequence as a motif.

\subsection{Connecting to Previous Studies}
Neural networks have produced state-of-the-art results on several important benchmark tasks related to genomic sequence classification \cite{alipanahi2015predicting,zhou2015predicting,quang2015danq}, making them a good candidate to use. However, \textit{why} these models work well has been poorly understood. Recent works have attempted to uncover the properties of these models, in which most of the work has been done on understanding image classifications using convolutional neural networks. Zeiler and Fergus \cite{zeiler2014visualizing} used a ``deconvolution'' approach to map hidden layer representations back to the input space for a specific example, showing the features of the image which were important for classification. Simonyan \etal \cite{simonyan2013deep} explored a similar approach by using a first-order Taylor expansion to linearly approximate the network and find the input features most relevant, and also tried optimizing image classes. Many similar techniques later followed to understand convolutional models \cite{mahendran2016visualizing, bach2015pixel}. Most importantly, researchers have found that CNNs are able to extract layers of translational-invariant feature maps, which may indicate why CNNs have been successfully used in genomic sequence predictions which are believed to be triggered by motifs.

On text-based tasks, there have been fewer visualization studies for DNNs. Karpathy \etal \cite{karpathy2015visualizing} explored the interpretability of RNNs for language modeling and found that there exist interpretable neurons which are able to focus on certain language structure such as quotes. Li \etal \cite{li2015visualizing} visualized how RNNs achieve compositionality in natural language for sentiment analysis by visualizing RNN embedding vectors as well as measuring the influence of input words on classification. Both studies show examples that can be validated by our understanding of natural language linguistics. Contrarily, we are interested in understanding DNA ``linguistics'' given DNNs (the opposite direction of Karpathy \etal \cite{karpathy2015visualizing} and Li \etal \cite{li2015visualizing}). 

The main difference between our work and previous works on images and natural language is that instead of trying to understand the DNNs given human understanding of such human perception tasks, we attempt to uncover critical signals in DNA sequences given our understanding of DNNs.

For TFBS prediction, Alipanahi \etal  \cite{alipanahi2015predicting} was the first to implement a visualization method on a DNN model. They visualize their CNN model by extracting motifs based on the input subsequence corresponding to the strongest activation location for each convolutional filter (which we call convolution activation). Since they only have one convolutional layer, it is trivial to map the activations back, but this method does not work as well with deeper models. We attempted this technique on our models and found that our approach using saliency maps outperforms it in finding motif patterns (details in section \ref{experiments}). Quang and Xie \cite{quang2015danq} use the same visualization method on their convolutional-recurrent model for noncoding variant prediction.

\section{Experiments and Results}
\label{experiments}
\subsection{Experimental Setup}
\paragraph{Dataset.\,}
In order to evaluate our DNN models and visualizations, we train and test on the 108 K562 cell ENCODE ChIP-Seq TF datasets used in Alipanahi \etal \cite{alipanahi2015predicting}. Each TF dataset has an average of 30,819 training sequences (with an even positive/negative split), and each sequence consists of 101 DNA-base characters (A,C,G,T). Every dataset has 1,000 testing sequences (with an even positive/negative split). Positive sequences are extracted from the hg19 genome centered at the reported ChIP-Seq peak. Negative sequences are generated by dinucleotide-preserving shuffle of the positive sequences. Due to the separate train/test data for each TF, we train a separate model for each individual TF dataset.

\paragraph{Variations of DNN Models.\,}
We implement several variations of each DNN architecture by varying hyperparameters. Table \ref{models} shows the different hyperparameters in each architecture. We trained many different hyperparameters for each architecture, but we show the best performing model for each type, surrounded by a larger and smaller version to show that it isn't underfitting or overfitting.

{\small
\begin{table}[tp]
\centering
\caption{Variations of DNN Model Hyperparameters}
{\begin{tabular}{|l|l|l|l|l|l|l|}
\hline
\textbf{Model} & \textbf{\begin{tabular}[c]{@{}l@{}}Conv. \\ Layers\end{tabular}} & \textbf{\begin{tabular}[c]{@{}l@{}}Conv. \\ Size ($n_{out}$)\end{tabular}} & \textbf{\begin{tabular}[c]{@{}l@{}}Conv. filter \\ Sizes ($k$)\end{tabular}} & \textbf{\begin{tabular}[c]{@{}l@{}}Conv. Pool\\ Size ($m$)\end{tabular}} & \textbf{\begin{tabular}[c]{@{}l@{}}LSTM\\ Layers\end{tabular}} & \textbf{\begin{tabular}[c]{@{}l@{}}LSTM\\ Size ($d$)\end{tabular}} \\ \hline
Small RNN    & N/A                                                              & N/A                                                            & N/A                                                                    & N/A                                                                & 1                                                              & 16
\\ \hline
Medium RNN    & N/A                                                              & N/A                                                            & N/A                                                                    & N/A                                                                & 1                                                              & 32                                                                 \\ \hline
Large RNN    & N/A                                                              & N/A                                                            & N/A                                                                    & N/A                                                                & 2                                                              & 32                                                                 \\ \hline
Small CNN      & 2                                                                & 64                                                             & 9,5                                                                    & 2                                                                  & N/A                                                            & N/A                                                                         \\ \hline
Medium CNN     & 3                                                                & 64                                                             & 9,5,3                                                                  & 2                                                                  & N/A                                                            & N/A                                                                          \\ \hline
Large CNN      & 4                                                                & 64                                                             & 9,5,3,3                                                                & 2                                                                  & N/A                                                            & N/A                                                                         \\ \hline
Small CNN-RNN  & 1                                                                & 64 & 5                                                                      & N/A                                                                  & 2 & 32                                                                
\\ \hline
Medium CNN-RNN  & 1                                                                & 128                                                            & 9                                                                      & N/A                                                                  & 1                                                              & 32                                                                 \\ \hline
Large CNN-RNN  & 2                                                                & 128                                                            & 9,5                                                                    & 2                                                                  & 1                                                              & 32                                                                 \\ \hline
\end{tabular}}
\label{models}
\end{table}
}

\paragraph{Baselines.\,} 
We use the ``MEME-ChIP \cite{machanick2011meme} sum'' results from Alipanahi \etal \cite{alipanahi2015predicting} as one prediction performance baseline. These results are from applying MEME-ChIP to the top 500 positive training sequences, deriving five PWMs, and scoring test sequences using the sum of scores using all five PWMs. We also compare against the CNN model proposed in Alipanahi \etal \cite{alipanahi2015predicting}. To evaluate motif extraction, we compare against the ``convolution activation'' method used in Alipanahi \etal \cite{alipanahi2015predicting} and Quang and Xie \cite{quang2015danq}, where we map the strongest first layer convolution filter activation back to the input sequence to find the most influential length-9 subsequence.

\subsection{TFBS Prediction Performance of DNN Models}
\label{performance}
Table \ref{auc_scores} shows the mean area under the ROC curve (AUC) scores for each of the tested models (from Table \ref{models}). As expected, the CNN models outperform the standard RNN models. This validates our hypothesis that positive binding sites are mainly triggered by local patterns or ``motifs'' that CNNs can easily find. Interestingly, the CNN-RNN achieves the best performance among the three deep architectures. To check the statistical significance of such comparisons, we apply a pairwise t-test using the AUC scores for each TF and report the two tailed p-values in Table \ref{p-values}. We apply the t-test on each of the best performing (based on AUC) models for each model type. All deep models are significantly better than the MEME baseline. The CNN is significantly better than the RNN and the CNN-RNN is significantly better than the CNN. In order to understand why the CNN-RNN performs the best, we turn to the dashboard visualizations.

\begin{table*}
\parbox{.65\linewidth}{
\centering
\caption{Mean AUC scores on the TFBS classification task}
{\begin{tabular}{|l|l|l|l|}
\hline
\textbf{Model} & \textbf{Mean AUC} & \textbf{Median AUC} & \textbf{STDEV} \\ \hline
MEME-ChIP \cite{machanick2011meme}        & 0.834             & 0.868               & 0.127          \\ \hline
DeepBind \cite{alipanahi2015predicting} (CNN)        & 0.903             & 0.931               & 0.091          \\ \hline
Small RNN    & 0.860             & 0.881               & 106          \\ \hline
Med RNN    & 0.876             & 0.905               & 0.116          \\ \hline
Large RNN    & 0.808             & 0.860               & 0.175          \\ \hline
Small CNN      & 0.896             & 0.918               & 0.098         \\ \hline
Med CNN        & 0.902             & 0.922               & 0.085          \\ \hline
Large CNN      & 0.880             & 0.890               & 0.093          \\ \hline
Small CNN-RNN  & 0.917             & 0.943              & 0.079          \\ \hline
Med CNN-RNN  & \textbf{0.925}             & \textbf{0.947}              & \textbf{0.073}          \\ \hline
Large CNN-RNN  & 0.918             & 0.944               & 0.081          \\ \hline
\end{tabular}}
\label{auc_scores}
}
\hfill
\parbox{.32\linewidth}{
\centering
\caption{AUC pairwise t-test}
{\begin{tabular}{|l|l|}
\hline
\textbf{Model Comparison\footnote{to compare models, we select the best performing model for each class}} & \textbf{p-value} \\ \hline
RNN vs MEME               & 5.15E-05           \\ \hline
CNN vs MEME               & 1.87E-19          \\ \hline
CNN-RNN vs MEME               & 4.84E-24          \\ \hline
CNN vs RNN                & 5.08E-04           \\ \hline
CNN-RNN vs RNN            & 7.99E-10           \\ \hline
CNN-RNN vs CNN            & 4.79E-22           \\ \hline
\end{tabular}}
\label{p-values}
}
\end{table*}

%To formally compare visualization methods, we evaluate each DNN's capability to find motifs, or consensus subsequences that significantly match to the corresponding JASPAR\cite{mathelier2015jaspar} motif for each TF using the Tomtom\cite{gupta2007quantifying} tool. Tomtom searches a query motif against a given motif database (and their reverse complements), and returns significant matches ranked by p-value indicating motif-motif similarity. Motif extraction for each of the models is implemented as follows. (1) From each positive test sequence (thus, 500 total for each TF dataset) we extract the saliency maps derived motif by selecting the contiguous length-9 subsequence that achieves the highest sum of contiguous length-9 saliency map values.  (2) For each positive test sequence, when extracting motifs using the temporal output scores, we select the length-9 subsequences that obtains the strongest score change from negative to positive output score. (3) Lastly, we compare each of the class-optimized sequences using the class optimization method against the JASPAR motifs. 

\subsection{Understanding DNNs Using the DeMo Dashboard}
\label{viz_evaluation}

\begin{figure}[tp]
\centering
\includegraphics[width=13cm]{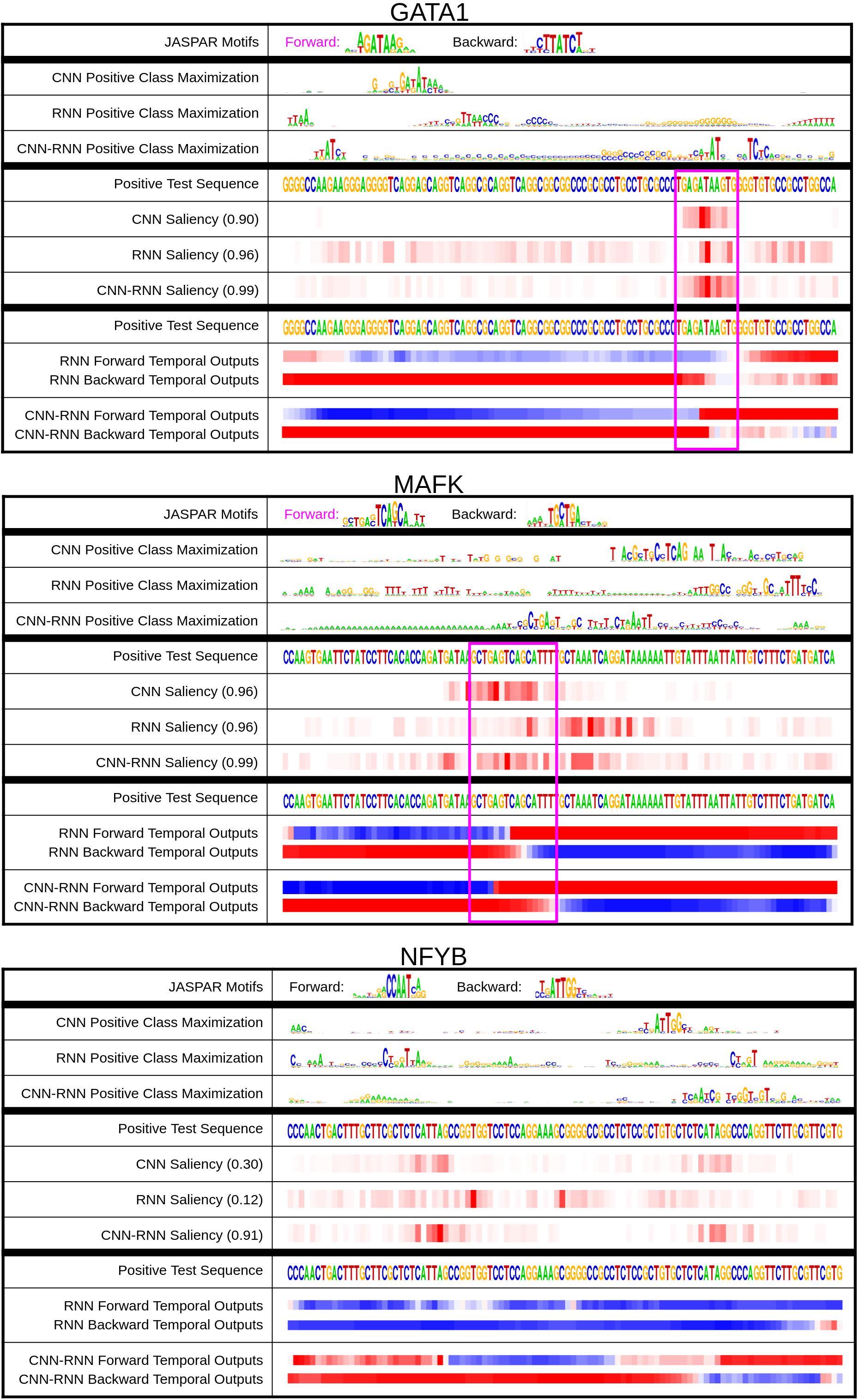}
\caption{\textbf{DeMo Dashboard}. Dashboard examples for GATA1, MAFK, and NFYB positive TFBS Sequences. The top section of the dashboard contains the Class Optimization (which does not pertain to a specific test sequence, but rather the class in general). The middle section contains the Saliency Maps for a specific positive test sequence, and the bottom section contains the temporal Output Scores for the same positive test sequence used in the saliency map. The very top contains known JASPAR motifs, which are highlighted by pink boxes in the test sequences if they contain motifs.}
\label{fig:dashboard}
\end{figure}

To evaluate the dashboard visualization methods, we first manually inspect the dashboard visualizations to look for interpretable signals. Figure \ref{fig:dashboard} shows examples of the DeMo Dashboard for three different TFs and positive TFBS sequences. We apply the visualizations on the best performing models of each of the three DNN architectures. Each dashboard snapshot is for a specific TF and contains (1) JASPAR\cite{mathelier2015jaspar} motifs for that TF, which are the ``gold standard'' motifs generated by biomedical researchers, (2) the positive TFBS class-optimized sequence for each architecture (for the given TF of interest), (3) the positive TFBS test sequence of interest, where the JASPAR motifs in the test sequences are highlighted using a pink box, (4) the saliency map from each DNN model on the test sequence, and (5) forward and backward temporal output scores from the recurrent architectures on the test sequence. In the saliency maps, the more red a position is, the more influential it is for the prediction.  In the temporal outputs, blue indicates a negative TFBS prediction while red indicates positive. The saliency map and temporal output visualizations are on the same positive test sequence (as shown twice). The numbers next to the model names in the saliency map section indicate the score outputs of that DNN model on the specified test sequence.

\paragraph{Saliency Maps (middle section of dashboard).\,} By visual inspection, we can see from the saliency maps that CNNs tend to focus on short contiguous subsequences when predicting positive bindings. In other words, CNNs  clearly model ``motifs'' that are the most influential for prediction. The  saliency maps of RNNs tend to be spread out more across the entire sequence, indicating that they focus on all nucleotides together, and infer relationships among them. The CNN-RNNs have strong saliency map values around motifs, but we can also see that there are other nucleotides further away from the motifs that are influential for the model's prediction. For example, the CNN-RNN model is 99\% confident in its GATA1 TFBS prediction, but the prediction is also influenced by nucleotides outside the motif. In the MAFK saliency maps, we can see that the CNN-RNN and RNN focus on a very wide range of nucleotides to make their predictions, and the RNN doesn't even focus on the known JASPAR motif to make its high confidence prediction.

\begin{table}
\centering
\caption{JASPAR motif matches against DeMo Dashboard and baseline motif finding methods using Tomtom}
{\begin{tabular}{p{2.1cm}|p{2.2cm}|p{3.7cm}|p{2.8cm}|p{3.1cm}|}
\cline{2-5}
                                       & \textbf{Saliency Map (out of 500)} & \textbf{Conv. Activations\cite{alipanahi2015predicting,quang2015danq} (out of 500)} & \textbf{Temporal Output (out of 500)} & \textbf{Class Optimization (out of 57)} \\ \hline
\multicolumn{1}{|l|}{\textbf{CNN}}     & 243.9                 & 173.4                     & N/A                     & 19                          \\ \hline
\multicolumn{1}{|l|}{\textbf{RNN}}     & 138.6                  & N/A                       & 53.5                      & 11                          \\ \hline
\multicolumn{1}{|l|}{\textbf{CNN-RNN}} &   168.1              & 74.2                     & 113.2                    & 13                          \\ \hline
\end{tabular}}
\label{motif-matches}
\end{table}
\setlength{\textfloatsep}{10pt}

\paragraph{Temporal Output Scores (bottom section of dashboard).\,}
For most of the sequences that we tested, the positions that trigger the model to switch from a negative TFBS prediction to positive are near the JASPAR motifs. We did not observe clear differences between the forward and backward temporal output patterns. 

In certain cases, it's interesting to look at the temporal output scores and saliency maps together. An important case study from our examples is the NFYB example, where the CNN and RNN perform poorly, but the CNN-RNN makes the correct prediction. We observe that the CNN-RNN is able to switch its classification from negative to positive, while the RNN never does. To understand why this may have happened, we can see from the saliency maps that the CNN-RNN focuses on two distinct regions, one of which is where it flips its classification from negative to positive. However, the RNN doesn't focus on either of the same areas, and may be the reason why it's never able to classify it as a positive sequence. The fact that the CNN is not able to classify it as a positive sequence, but focuses on the same regions as the CNN-RNN (from the saliency map), may indicate that it is the temporal dependencies between these regions which influence the binding. In addition, the fact that there is no clear JASPAR motif in this sequence may show that the traditional motif approach is not always the best way to model TFBSs.

\paragraph{Class Optimization (top section of dashboard).\,}
Class optimization on the CNN model generates concise representations which often resemble the known motifs for that particular TF. For the recurrent models, the TFBS positive optimizations are less clear, though some aspects  stand out (like ``AT'' followed by ``TC'' in the GATA1 TF for the CNN-RNN). We notice that for certain DNN models, their class optimized sequences optimize the reverse complement motif (e.g. NFYB CNN optimization). The class optimizations can be useful for getting a general idea of what triggers a positive TFBS for a certain TF.

\paragraph{Automatic Motif Extraction from Dashboard.\,}

In order to evaluate each DNN's capability to automatically extract motifs, we compare the found motifs of each method (introduced in section \ref{e2emotif}) to the corresponding JASPAR motif, for the TF of interest. We do the comparison using the Tomtom\cite{gupta2007quantifying} tool, which searches a query motif against a given motif database (and their reverse complements), and returns significant matches ranked by p-value indicating motif-motif similarity. Table \ref{motif-matches} summarizes the motif matching results comparing visualization-derived motifs against known motifs in the JASPAR database. We are limited to a comparison of 57 out of our 108 TF datasets by the TFs which JASPAR has motifs for. We compare four visualization approaches: Saliency Map, Convolution Activation\cite{alipanahi2015predicting,quang2015danq}, Temporal Output Scores and Class Optimizations. The first three techniques are sequence specific, therefore we report the average number of motif matches out of 500 positive sequences (then averaged across 57 TF datasets). The last technique is for a particular TFBS positive class.  %The fact that the CNN is better than the other two at finding known motifs indicates that CNNs clearly are highly influenced by motifs. 

We can see from Table \ref{motif-matches} that across multiple visualization techniques, the CNN finds motifs the best, followed by the CNN-RNN and the RNN. However, since CNNs perform worse than CNN-RNNs by AUC scores, we hypothesize that this demonstrates that it is also important to model sequential interactions among motifs. In the CNN-RNN combination, CNN acts like a ``motif finder'' and the RNN finds dependencies among motifs. This analysis shows that visualizing the DNN classifications can lead to a better understanding of DNNs for TFBSs.

\section{Conclusions and Future Work}
Deep neural networks (DNNs) have shown to be the most accurate models for TFBS classification. However, DNN models are hard to interpret, and thus their adaptation in practice is slow. In this work, we propose the Deep Motif (DeMo) Dashboard to explore three different DNN architectures on TFBS prediction, and introduce three visualization methods to shed light on how these models work. Although our visualization methods still require a human practitioner to examine the dashboard, it is a start to understand these models and we hope that this work will invoke further studies on visualizing and understanding DNN based genomic sequences analysis. Furthermore, DNN models have recently shown to provide excellent results for epigenomic analysis \cite{Singh01092016}. We plan to extend our DeMo Dashboard to related applications.

\bibliographystyle{plain}
\bibliography{deep_gdashboard}

\end{document}